\PassOptionsToPackage{warn}{textcomp}
\documentclass[sigconf]{acmart}
\pdfoutput=1
\usepackage{savesym}
\usepackage[russian,english]{babel}

\usepackage[section]{placeins}
\usepackage{color, colortbl}
\usepackage{times}
\usepackage{latexsym}
\usepackage[nomessages]{fp}
\usepackage{multirow}

\usepackage{subcaption}

\usepackage{contour}
\usepackage[normalem]{ulem}

\contourlength{0.8pt}
\newcommand{\myuline}[1]{%
  \uline{\phantom{#1}}%
  \llap{\contour{white}{#1}}%
}

\usepackage{graphicx}
\usepackage{xcolor}
\usepackage{url}
\usepackage{soul}
\usepackage[T2A,T1]{fontenc}
\usepackage[utf8]{inputenc}
\usepackage{csquotes}
\usepackage{amsmath}
\usepackage{amssymb}
\usepackage{textcomp}
\usepackage{tabularx}
\usepackage{fixltx2e}
\mathchardef\mhyphen="2D

\AtBeginDocument{%
  \providecommand\BibTeX{{%
    \normalfont B\kern-0.5em{\scshape i\kern-0.25em b}\kern-0.8em\TeX}}}

\copyrightyear{2020}
\acmYear{2020}
\setcopyright{acmcopyright}
\acmPrice{15.00}
\acmDOI{10.1145/nnnnnnn.nnnnnnn}
\acmISBN{978-1-4503-XXXX-X/20/06}
\acmPrice{15.00}

\acmConference[WIMS '20]{WIMS '20}{June 30th - July 3rd, 2020}{Biarritz, France}

\begin{document}

\title{Attention-Based Neural Networks for Sentiment Attitude Extraction using Distant Supervision}

\author{Nicolay Rusnachenko}
\email{kolyarus@yandex.ru}
\affiliation{%
    \institution{Bauman Moscow State Technical University}
    \city{Moscow}
    \country{Russia}
}

\author{Natalia Loukachevitch}
\email{louk\_nat@mail.ru}
\affiliation{%
  \institution{Lomonosov Moscow State University}
  \city{Moscow}
  \country{Russia}
}

\renewcommand{\shortauthors}{N. Rusnachenko and N. Loukachevitch}

\newcommand{\rusentrel}{RuSentRel}
\newcommand{\ruattitudes}{RuAttitudes}
\newcommand{\rusentiframes}{RuSentiFrames}
\newcommand{\rusentilex}{RuSentiLex}

\newcommand{\dsBaseImproveRange}{10}
\newcommand{\dsAttImproveRange}{3}
\newcommand{\useDS}{$\bullet$}

\newcommand{\test}{\textsc{test}}
\newcommand{\train}{\textsc{train}}
\newcommand{\testSc}{\textsubscript{\test{}}}
\newcommand{\trainSc}{\textsubscript{\train{}}}

\newcommand{\epochsToTest}{10}
\newcommand{\epochsCount}{150}
\newcommand{\pcnnIncrease}{1.4}
\newcommand{\ianIncrease}{1.2}
\newcommand{\bilstmIncrease}{5.9}

\newcommand{\raEntity}[1]{\myuline{#1}$_e$}
\newcommand{\raFrame}[2]{\textit{\textbf{{#1}$_{{#2}}$}}}
\newcommand{\raAtt}[2]{\texttt{#1}$\to$\texttt{#2}}

\newcommand{\maskE}{$\myuline{E}$}
\newcommand{\maskEObj}{$\myuline{E}_{obj}$}
\newcommand{\maskESubj}{$\myuline{E}_{subj}$}

\newcommand{\argZero}{\texttt{Arg0}}
\newcommand{\argOne}{\texttt{Arg1}}
\newcommand{\polarity}{\texttt{A0}$\to$\texttt{A1}}
\newcommand{\mpolarity}{\texttt{A0} \to \texttt{A1}}

\newcommand{\featuresCount}{k}
\newcommand{\embSize}{m}
\newcommand{\ctxSize}{n}            
\newcommand{\pairDist}{\eta}        
\newcommand{\labelsCount}{c}

\newcommand{\word}[1]{t_{#1}}
\newcommand{\embWord}[1]{x_{#1}}
\newcommand{\embFeature}[1]{f_{#1}}
\newcommand{\hidden}[1]{h_{#1}}
\newcommand{\entity}[1]{e_{#1}}

\newcommand{\featureSet}{F}
\newcommand{\embSet}{X}

\newcommand{\sentEmbSize}{|\sentenceEmbedding{}|}

\newcommand{\entityGroup}{\textsc{entities}}
\newcommand{\wordsGroup}{\textsc{words}}

\newcommand{\han}{\textsc{HAN}}
\newcommand{\lstm}{\textsc{LSTM}}

\newcommand{\Any}{*}
\newcommand{\no}{$\cdot$}
\newcommand{\dd}[1]{\Delta_{#1}}

\newcommand{\fm}[1]{F1_{#1}}
\newcommand{\fmCv}[1]{F1_{cv}^{#1}}

\newcommand{\dsm}{\textsc{DS}}
\newcommand{\slm}{\textsc{SL}}

\newcommand{\blstm}{\textsc{B}\lstm{}}
\newcommand{\bilstm}{\textsc{B}i\lstm{}}
\newcommand{\bilstmPZhou}{\textsc{Att}-\blstm{}}
\newcommand{\bilstmZYang}{\textsc{Att}-\blstm{}$^{z\mhyphen yang}$}

\newcommand{\featureEnds}{${{att\mhyphen ends}}$}
\newcommand{\featureFrames}{${{att\mhyphen frames}}$}
\newcommand{\featureEf}{${{att\mhyphen ef}}$}

\newcommand{\attCnn}{\textsc{AttCNN}}

\newcommand{\ian}{\textsc{IAN}}
\newcommand{\ianEnds}{\ian{}$_{{ends}}$}
\newcommand{\ianFrames}{\ian{}$_{{frames}}$}
\newcommand{\ianSef}{\ian{}$_{{sef}}$}
\newcommand{\ianSe}{\ian{}$_{{se}}$}
\newcommand{\ianEf}{\ian{}$_{{ef}}$}

\newcommand{\pcnn}{\textsc{PCNN}}
\newcommand{\pcnnEnds}{\textsc{Att}\pcnn{}$_e$}
\newcommand{\pcnnFrames}{\textsc{Att}\pcnn{}$_f$}
\newcommand{\pcnnEf}{\textsc{Att}\pcnn{}$_{ef}$}

\newcommand{\cnn}{\textsc{CNN}}
\newcommand{\cnnEnds}{\textsc{Att}\cnn{}$_e$}
\newcommand{\cnnFrames}{\textsc{Att}\cnn{}$_f$}
\newcommand{\cnnEf}{\textsc{Att}\cnn{}$_{ef}$}

\newcommand{\attCnnAny}{\textsc{Att}\cnn{}$_{*}$}
\newcommand{\attPcnnAny}{\textsc{Att}\pcnn{}$_{*}$}

\newcommand{\ianHidden}{\textbf{h}}

\newcommand{\devDatasetShort}{RA}

\newcommand{\twoScale}{\textsc{Two-scale}}
\newcommand{\threeScale}{\textsc{Three-scale}}

\newcommand{\tokenGroup}{\textsc{tokens}}
\newcommand{\prepGroup}{\textsc{prep}}
\newcommand{\framesGroup}{\textsc{frames}}
\newcommand{\sentGroup}{\textsc{sentiment}}

\newcommand{\precision}{$P$}
\newcommand{\recall}{$R$}
\newcommand{\fmeasure}{$F1$}

\newcommand{\stemmer}{\url{https://tech.yandex.ru/mystem/}}

\newcommand{\rusentrelLink}{\url{https://github.com/nicolay-r/RuSentRel/tree/v1.1}}
\newcommand{\ruSentiFramesLink}{\url{https://github.com/nicolay-r/RuSentiFrames/tree/v1.0}}
\newcommand{\devDatasetLink}{\url{https://github.com/nicolay-r/RuAttitudes/tree/v1.0}}
\newcommand{\experimentsLink}{\url{https://github.com/nicolay-r/attitudes-extraction-ds}}

\newcommand{\bagsCount}{l}
\newcommand{\sentencesCount}{t}
\newcommand{\classesCount}{c}
\newcommand{\sentenceEmbedding}{s}
\newcommand{\sentenceEmbeddingSet}{E_{s}}

\begin{abstract}

In the sentiment attitude extraction task, the aim is to identify
<<attitudes>> -- sentiment relations between entities mentioned 
in text. 
In this paper, we provide a study on attention-based context encoders in the
sentiment attitude extraction task.
For this task, we adapt attentive context encoders of two types:
(1)~feature-based;
(2)~self-based.
In our study, we utilize the corpus of Russian analytical texts 
\rusentrel{} and automatically constructed news collection  \ruattitudes{} for enriching the training set.
We consider the problem of attitude extraction as two-class (positive, negative) and three-class
(positive, negative, neutral) classification tasks for whole documents.
Our experiments\footnote{\url{https://github.com/nicolay-r/attitude-extraction-with-attention-and-ds}} with the \rusentrel{} corpus show that
the three-class classification models, which employ the  \ruattitudes{} corpus  for training, result in 10\%
increase and extra 3\% by \fmeasure{}, when model architectures include the attention mechanism.
We also provide the analysis of attention weight distributions in dependence on the term type.

\end{abstract}


\maketitle

\section{Introduction}

    Classifying relations between entities mentioned in texts remains one of the difficult tasks in natural language processing (NLP).
    The sentiment attitude extraction  aims to 
    seek for positive/negative relations between  objects expressed as named entities  in texts~\cite{rusnachenko2018neural}. 
    For example, in Figure~\ref{fig:intro_example} 
      named entities <<Russia>> and <<NATO>> have the  negative attitude towards each other
    with additional indication of other named entities.
    \begin{table}
    \begin{center}
        \begin{tabular}{|c|p{6.2cm}|}
            \hline
            \multicolumn{1}{|c|}{\textsc{context}} & \begin{otherlanguage*}{russian}
                При этом \myuline{Москва} неоднократно подчеркивала, что ее активность на \myuline{балтике}
                является ответом именно на действия \textbf{\myuline{НАТО}} и эскалацию враждебного подхода к
                \textbf{\myuline{России}} вблизи ее восточных границ $\ldots$
            \end{otherlanguage*}
            \\ \cline{2-2}
                                                   & Meanwhile \myuline{Moscow} has repeatedly emphasized that its activity in the
                                                     \myuline{Baltic Sea} is a response precisely to actions of \textbf{\myuline{NATO}}
                                                     and the escalation of the hostile approach to \textbf{\myuline{Russia}} near its eastern borders $\ldots$
            \\ \hline
            \multicolumn{1}{|c|}{\textsc{attitudes}} & \texttt{NATO}$\to$\texttt{Russia}: neg \\
                                                    & \texttt{Russia}$\to$\texttt{NATO}: neg \\ \hline
        \end{tabular}
    \end{center}
    \captionof{figure}{
    Example of a context with  attitudes mentioned in it;
    named entities <<Russia>> and <<NATO>> have the negative attitude towards each other
    with additional indication of other named entities.}

    \label{fig:intro_example}
\end{table}

When extracting relations from  texts, one encounters the complexity of the sentence structure; sentences can contain many named entity mentions;  a single opinion might comprise several sentences.

    This paper is devoted to study  of models for targeted sentiment analysis with attention.
    The intuition exploited in the models with attentive encoders  is that only some terms in the context are relevant for attitude indication.
    The interactions of words, not just their isolated presence, may reveal the specificity of
    contexts with attitudes of different polarities.
    We additionally used the distant supervision (DS)~\cite{mintz2009distant} technique to fine-tune the attention mechanism
    by providing  relevant contexts, with words that indicate the presence of attitude.
    Our contribution in this paper is three-fold:
    \begin{itemize}
        \item We apply attentive encoders based on
        (1) attitude participants and
        (2) context itself;

        \item We conduct the experiments on the \rusentrel{}~\cite{loukachevitch2016creating} collection
        using the distant supervision technique in the  training process.
        The results demonstrate that the application of attention-based encoders
        enhance  quality by \dsAttImproveRange{}\% \fmeasure{} in the three-class classification task;

        \item We provide an analysis of weight distribution to illustrate the influence of distant supervision onto informative terms selection.
    \end{itemize}

\section{Related Work}
\label{sec:review}

In previous works, various neural network approaches for targeted sentiment analysis were proposed.
In~\cite{rusnachenko2018neural} the authors utilize 
convolutional neural networks (\cnn{}).
Considering  relation extraction as a three-scale classification task of contexts with attitudes in it,
the authors subdivide each context into \textit{outer} and \textit{inner} 
(relative to attitude participants) to apply Piecewise-\cnn{} (\pcnn{})~\cite{zeng2015distant}.
The latter architecture utilizes a specific idea of the \textit{max-pooling} operation.
Initially, this is an operation, which  extracts 
the maximal values within each convolution.
However, for relation classification, it reduces information extremely rapid and blurs 
significant aspects of 
context parts. In case of PCNN, separate max-pooling operations are applied to outer and inner contexts.
In the experiments, the authors revealed a fast training process and a slight improvement in the \pcnn{} results  in comparison to \cnn{}.

In~\cite{shen-huang-2016-attention}, the authors proposed an attention-based CNN model for semantic relation classification~\cite{hendrickx2009semeval}.
The authors utilized the attention mechanism to select the most relevant context words with respect to participants of a  semantic relation.
The architecture of the attention model is a multilayer perceptron (MLP), which  calculates the weight of a word in context with respect to the entity.
The resulting \textsc{AttCNN} model outperformed several \textsc{CNN} and \textsc{LSTM} based approaches with $2.6\mhyphen{}3.8\%$ by F1-measure.

In \cite{zhou2016attention,yang2016hierarchical}, the authors experimented with self-based attention models, in which \textit{targets} became adapted automatically during the training process. 
The authors considered the attention as context word quantification with respect to abstract targets.
In \cite{yang2016hierarchical}, the authors brought a similar idea also onto the sentence level. 
The obtained hierarchical model was called as \han{}.


In~\cite{rusnachenko2019distant}, authors apply distant supervision (DS) approach
to developing an automatic collection for the sentiment attitude extraction task in the news domain.
A combination of two labeling methods (1)~pair-based and (2)~frame-based were used
to perform context labeling.
The developed collection was called as~\ruattitudes{}.
Experimenting with the \rusentrel{} corpus, the authors consider the problem of sentiment attitude extraction as a two-class classification task
and mention the 13.4\% increase by \fmeasure{} when models trained with an application of~\ruattitudes{}
over models which training relies on supervised learning.

For Russian, Archipenko et al.~\cite{arkhipenko2016comparison} compared neural architectures
for entity-related tweet setiment classification; they found that the best results were obtained with the GRU neural model~\cite{cho2014learning}.
The authors of~\cite{rogers2018rusentiment} annotated more than $31$~thousand social media posts in Russian
with three sentiment categories and compared several baseline classification methods, obtaining the best
results with a four-layer neural model with non-linear activations between layers.
These results were improved in~\cite{kuratov2019adaptation}, where the authors applied the BERT model
trained on Russian data (RuBERT).
Tutubalina et al.~\cite{tutubalina2020russian} compared several neural network models to extract positive or
negative adverse drug reactions in Russian social network texts.

\section{Resources}
\label{sec:data}

In our study we utilize the following collections:
(1) \rusentrel{} as a source of news texts with manually provided attitude labeling in it, and
(2) automatically developed \ruattitudes{} collection, which addresses the lack of training examples in \rusentrel{}.

We also use two Russian sentiment resources: the \rusentilex{} lexicon~\cite{loukachevitch2016creating},
which contains words and expressions of the Russian language with sentiment labels and the \rusentiframes{} lexicon~\cite{rusnachenko2019distant},
which provides several types of sentiment attitudes for situations associated with specific Russian predicates.

\subsection{\rusentrel{} collection}
\label{sec:rusentrel}

We consider sentiment analysis of Russian analytical articles collected in the \rusentrel{} corpus~\cite{loukachevitch2018extracting}.
The corpus comprises texts in the international politics domain and contains a lot of opinions.
The articles are labeled with  annotations of two types:
(1) the author's opinion on the subject matter of the article;
(2) the attitudes between the participants of the described situations.
The annotation of the latter type includes 2000 relations across 73 large analytical texts.
Annotated sentiments can be only \textit{positive} or \textit{negative}.
Additionally,  each text is provided  with annotation of mentioned named entities.
Synonyms and variants of named entities are also given, which allows not to deal with the coreference of named entities.

\subsection{\rusentiframes{} lexicon}
\label{sec:rusentiframes}

The \rusentiframes{}\footnote{\ruSentiFramesLink{}} lexicon describes sentiments and connotations conveyed with a predicate in a verbal or nominal form  \cite{rusnachenko2019distant}, such as \begin{otherlanguage*}{russian}"осудить, улучшить, преувеличить" \end{otherlanguage*}
(to condemn, to improve, to exaggerate), etc.
The structure of the frames in RuSentFrames comprises:
(1) the set of predicate-specific roles;
(2) frames dimensions such as the attitude of the author towards participants of the situation, attitudes between the participants, effects for participants. Currently, RuSentiFrames contains frames for more than 6 thousand  words and expressions.

\begin{table}[!htp]
    \begin{tabular}{ll}
        \hline
        Frame              & \begin{otherlanguage*}{russian} "Одобрить" \end{otherlanguage*} (Approve) \\ \hline
        \textsc{roles}     & \texttt{A0}: who approves               \\
                           & \texttt{A1}: what is approved           \\ \hline
        \textsc{polarity}  & \texttt{A0} $\to$ \texttt{A1}, pos, 1.0 \\
                           & \texttt{A1} $\to$ \texttt{A0}, pos, 0.7 \\ \hline
        \textsc{effect}    & \texttt{A1}, pos, 1.0                   \\ \hline
        \textsc{state}     & \texttt{A0}, pos, 1.0                   \\
                           & \texttt{A1}, pos, 1.0                   \\ \hline
    \end{tabular}
    \caption{Example description of frame \protect\begin{otherlanguage*}{russian} <<Одобрить>> \protect\end{otherlanguage*} (Approve) in RuSentiLex lexicon.}
    \label{tab:frames-example}
\end{table}

In RuSentiFrames, individual  semantic roles are numbered, beginning with zero. For a particular predicate entry, \argZero{} is generally the argument exhibiting features of a Prototypical Agent, while \argOne{} is a Prototypical Patient or Theme \cite{dowty1991thematic}.
In the main part of the frame, the most applicable for the current study is 
the polarity of \argZero{} with a respect to \argOne{} (\polarity{}).
Table~\ref{tab:frames-example} provides an example of frame
\begin{otherlanguage*}{russian} "одобрить" \end{otherlanguage*} (to approve).

\subsection{\ruattitudes{}}
\label{sec:ruattitudes}

The \ruattitudes{}~\cite{rusnachenko2019distant} is a corpus of  news texts automatically  labeled using distant supervision approach.
These are news stories from specialized political sites and Russian sites of world-known news agencies published in 2017.
The news texts are annotated with attitudes between participants, which sentiments can be only positive or negative.
In comparison with \rusentrel{}, the \ruattitudes{} corpus includes $14.6$~K attitudes gathered across $13.4$~K news texts.

Every news text is presented as a sequence of its contexts, where the first context is a news \textit{title} and others are news
content or \textit{sentences}.
For a particular news story, the \ruattitudes{} corpus keeps information of only those contexts, which has at least one attitude mentioned in it.
Each context is presented as a sequence of words with named entities markup.
According to Section~\ref{sec:review}, the authors considered an application of two factors (1) \textsc{Pair-based} and (2) \textsc{Frame-based}
in order to define the fact of presence and sentiment polarity of an \textit{attitude}, which is described by a pair of mentioned named entities.

\begin{table}
    \begin{center}
        \begin{tabular}{|p{8.1cm}|}
            \hline
            \multicolumn{1}{|l|}{\textsc{title}} \\ \hline
            \begin{otherlanguage*}{russian}
                Маккейн: \raEntity{США} \raFrame{продолжат}{pos} \raFrame{поддержку}{pos} \raEntity{Грузии}
            \end{otherlanguage*}
            \\
            McCain: \raEntity{USA} \raFrame{continue}{pos} \raFrame{supporting}{pos} \raEntity{Georgia}
            \\ \hline

            \multicolumn{1}{c}{$\downarrow$ \raAtt{USA}{Georgia}$_{pos}$} \\ \hline

            \textsc{sentence}: 5 \\ \hline
            \begin{otherlanguage*}{russian}
            <<\raEntity{США} и далее \raFrame{продолжат}{pos} \raFrame{поддержку}{pos} свободы, суверенитета и территориальной
                целостности \raEntity{Грузии} в рамках международно признанных границ страны>>, -- сказал он.
            \end{otherlanguage*} \\

            <<\raEntity{USA} and in further \raFrame{continue}{pos} \raFrame{support}{pos} freedom, sovereignty and territorial integrity
            \raEntity{Georgia} within the internationally recognized borders of the country>>, -- he said. \\

            \hline

            \multicolumn{1}{c}{$\downarrow$ \raAtt{USA}{Georgia}$_{pos}$} \\ 
            
            \hline

            \textsc{sentence}: 11 \\ \hline
            \begin{otherlanguage*}{russian}
                29 декабря премьер-министр \raEntity{Квирикашвили} сообщил, что правительство \raEntity{Грузии}
                установило первые контакты с новой администрацией \raEntity{США}.
            \end{otherlanguage*} \\
            29'th december prime-minister \raEntity{Kvirikashvili} reported, that the government of \raEntity{Georgia}
            has established first contacts with the new \raEntity{USA} administration. \\ \hline

        \end{tabular}

    \end{center}

    \captionof{figure}{Example of news (\#11323) description from \ruattitudes{}-1.1 collection illustrates the attitude \raAtt{USA}{Georgia}$_{pos}$ which is 
    annotated by \textsc{Frame-based} and \textsc{Pair-based} factors in news title with 
    the corresponding appearance of $\left<USA, Georgia\right>$ pair in the sentences (\#5, \#11) of news content.}

    \label{fig:ruattitudes-example}
\end{table}

\textsc{Pair-based} factor assumes to perform annotation using a list of entity pairs with preassigned sentiment polarities.
In turn, \textsc{Frame-based} factor utilizes infomation from the~\rusentiframes{} lexicon (Section~\ref{sec:rusentiframes}) in order to perform annotation.
The context is retrieved in case when both factors are met.
Due to the latter, it is worth to mention the specifics of the \textsc{Frame-based} factor.
A pair of neighbour named entities is considered as having a sentiment attitude when a news title has the following structure:
\begin{center}
    \raEntity{Subject} \hspace{0.1cm} $\ldots$ \hspace{0.1cm} \{\raFrame{frame}{\mpolarity{}}\}$_k$ \hspace{0.1cm} $\ldots$ \hspace{0.1cm} \raEntity{Object}
    \hspace{0.5cm}
\end{center}
where $k$ corresponds to the size of the non-empty set.
The sentiment score is considered \textit{positive} in the case when all the frame entries of the set are equally positive in terms of \polarity{} polarity values.
Otherwise, the sentiment is considered \textit{negative}.
The annotated attitude is then utilized in news content filtering.
Sentences that has no subject and object entries of the related attitude are discarded.
Figure~\ref{fig:ruattitudes-example} provides an example of a news text,
in which attitude $\left<Georgia, USA\right>$ assumes to be annotated by \textsc{Frame-based} factor as positive:
all the frames mentioned between attitude ends (to continue, to support) conveys the same positive sentiment value of \polarity{} polarity.

%
%
%

\begin{figure}[t]
    \includegraphics[width=\linewidth]{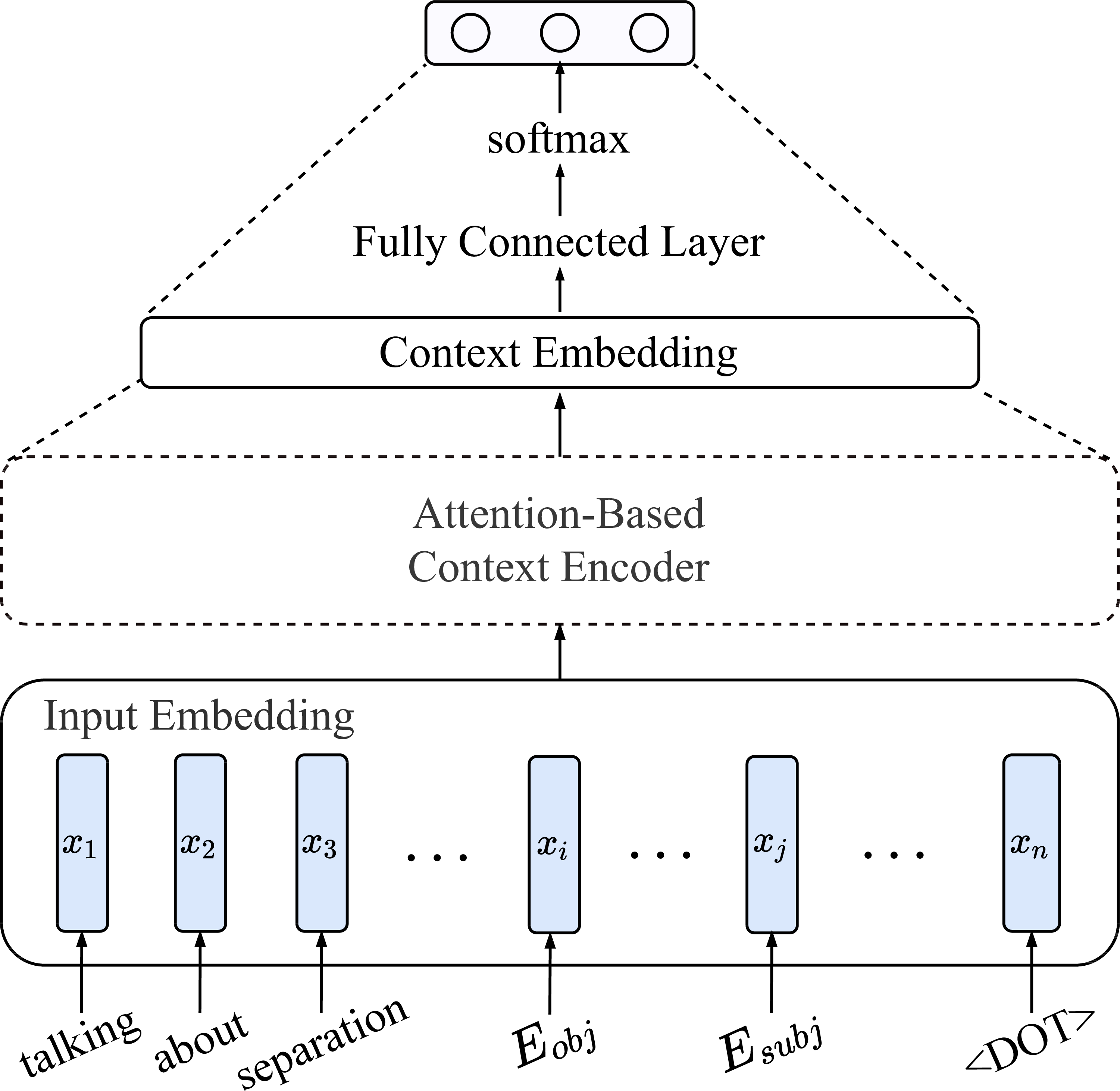}~\vspace{0.2cm}
    \caption{
     General, context-based 3-scale (positive, negative, neutral) classification model,
     with details on <<Attention-Based Context Encoder>> block in Section~\ref{sec:mlp-based}~and~\ref{sec:self-based}.
    }
    \label{fig:model}
\end{figure}

\section{Model}
\label{sec:model}

In this paper, the problem of sentiment attitude extraction is treated as a classification task of two types:
two-scale and three-scale.
Given a pair of named entities, we predict a sentiment label of
a pair, which could be as follows:
\begin{itemize}
    \item sentiment, i.e. positive or negative (two-scale classification format);
    \item sentiment or \textit{neutral}.
\end{itemize}

As the \rusentrel{} corpus provides opinions with positive or negative sentiment labels only~(Section~\ref{sec:data}),
we automatically added neutral sentiments for all pairs not
mentioned in the annotation and co-occurred in the same sentences of the collection texts.

\newcommand{\fframe}[1]{\textbf{\textit{#1}}}

\begin{table}[t]
    \begin{center}
    \begin{tabular}{|p{8.1cm}|}
        \hline
        \multicolumn{1}{|c|}{\textsc{context}} \\ \hline
        \begin{otherlanguage*}{russian}
        Говорить о разделении \myuline{кавказского региона} из-за конфронтации
        \textbf{\myuline{России}$_{obj}$} и
        \textbf{\myuline{Турции}$_{subj}$}
        пока не приходится, хотя опасность есть.
        \end{otherlanguage*}  \\  \hline
        Talking about the separation of the
        \myuline{Caucasus region} due to the confrontation between
        \textbf{\myuline{Russia}$_{obj}$} and
        \textbf{\myuline{Turkey}$_{subj}$}
        is not necessary, although there is a danger. \\ \hline
        \multicolumn{1}{c}{\large $\downarrow$} \\ \hline
        \multicolumn{1}{|c|}{\textsc{terms}} \\ \hline
        Talking about the separation of the \maskE{} due to the
        \fframe{confrontation$_{neg}$}
        between \maskEObj{} and \maskESubj{} is
        \fframe{not-necessary$_{neg}$}
        \texttt{<COMMA>} although there is a danger \texttt{<DOT>} \\
        \hline
    \end{tabular}
    \end{center}
    \captionof{figure}{
        An example of a context processing into a sequence of terms;
        attitude participants $\left<Russia, Turkey\right>$ and other mentioned entities become masked;
        frames are italic and optionally suffixed with the sentiment value of
        \polarity{} polarity.
    }
    \label{fig:processing-example}
\end{table}

We consider a \textit{context} as a text fragment that is limited by a single sentence and includes a pair of named
entities.
The general architecture is presented in Figure~\ref{fig:model}, where
the sentiment could be extracted from the context.
To present a context, we treat the original text as a sequence of terms
$[\word{1}, \ldots, \word{\ctxSize{}}]$ limited by $\ctxSize{}$,
with the distance between attitude participants limited by $\pairDist{}$ terms.
Each term belongs to one of the following groups:
\entityGroup{},
\framesGroup{}, \tokenGroup{}, and
\wordsGroup{} (if none of the prior has not been matched).

We use masked representation for attitude participants (\maskEObj{}, \maskESubj{})
and mentioned named entities (\maskE{}) to prevent models from capturing related information.
To represent \framesGroup{}, we combine 
a frame entry with the corresponding \polarity{} sentiment polarity value
(and \textit{neutral} if the latter is absent).
We also invert sentiment polarity when an entry has
\begin{otherlanguage*}{russian}"не"\end{otherlanguage*}
(not) preposition.
The~\tokenGroup{} group includes: punctuation marks, numbers, url-links. 
Each term of \wordsGroup{} is considered in a lemmatized\footnote{\stemmer{}} form.

Figure~\ref{fig:processing-example} provides an example of a context processing into a sequence of input terms.
All entries are encoded with the negative polarity \polarity{}:
\begin{otherlanguage*}{russian}"конфронтация"\end{otherlanguage*} (confrontation)
has a negative polarity, and
\begin{otherlanguage*}{russian}"не приходится"\end{otherlanguage*} (not necessary)
has a positive polarity of entry "necessary" which is inverted due to the "not" preposition.

To represent the context in a model, each term is embedded with a 
vector of fixed dimension.
The sequence of embedded vectors $\embSet{} = [\embWord{1}, \ldots, \embWord{\ctxSize{}}]$ is
denoted as \textit{input embedding} ($\embWord{i} \in \mathbb{R}^\embSize{}, i \in \overline{1..\ctxSize{}}$).
Sections~\ref{sec:mlp-based}~and~\ref{sec:self-based} provide an encoder implementation in details.
In particular, each encoder relies on input embedding and generates
output \textit{embedded context} vector $\sentenceEmbedding{}$.

In order to determine a sentiment class by the embedded context $\sentenceEmbedding{}$, 
we apply:
(1)~the hyperbolic tangent activation function towards $\sentenceEmbedding{}$ and
(2)~transformation through the \textit{fully connected layer}:

\newcommand{\resultVector}{r}
\newcommand{\WHidden}{W_\resultVector{}}
\newcommand{\BHidden}{b_\resultVector{}}

\begin{equation}
    \resultVector{} = \WHidden{} \cdot \tanh(s) + \BHidden{}
    \label{eq:fc}
\end{equation}

\newcommand{\smProb}{\rho}
\newcommand{\softmaxArgIndex}{i}
\newcommand{\softmaxDimSize}{K}

In Formula~\ref{eq:fc}, $\WHidden{} \in \mathbb{R}^{\sentEmbSize{} \times \labelsCount{}}$ and
$\BHidden{} \in \mathbb{R}^\labelsCount{}$
correspond to the hidden states;
$\sentEmbSize{}$ correspond to the size of vector $\sentenceEmbedding{}$,
and $\labelsCount{} \in \{2, 3\}$ is a number of classes.
Finally, the result $o=\sigma(r, \labelsCount{})$ 
is an output vector of probabilities, which
is computed by:
\begin{equation}
    \sigma(z, \softmaxDimSize{})_\softmaxArgIndex{} = \frac{\exp(z_\softmaxArgIndex{})}{\sum_{j=1}^{\softmaxDimSize{}}{\exp(z_j)}}
    \hspace{1cm}
    z \in \mathbb{R}^{\softmaxDimSize{}}
    \label{eq:softmax}
\end{equation}
\section{Feature Attentive Context Encoders}
\label{sec:mlp-based}

In this section, we consider \textit{features} as a
significant for attitude identification context terms, towards which we would like
to quantify the relevance of each term in the context.
For a particular context, we select embedded values of the attitude participants (\maskEObj{}, \maskESubj{}).

\newcommand{\featEmb}{\hat{s}}
\newcommand{\mlpHidden}{\textbf{h}_\textsc{mlp}}
\newcommand{\convCount}{\textbf{c}}
\newcommand{\featureBasedVector}{s_{f}}
\newcommand{\cnnBasedVector}{s_{cnn}}

\begin{figure}[!t]
    \centering
    \includegraphics[width=\linewidth]{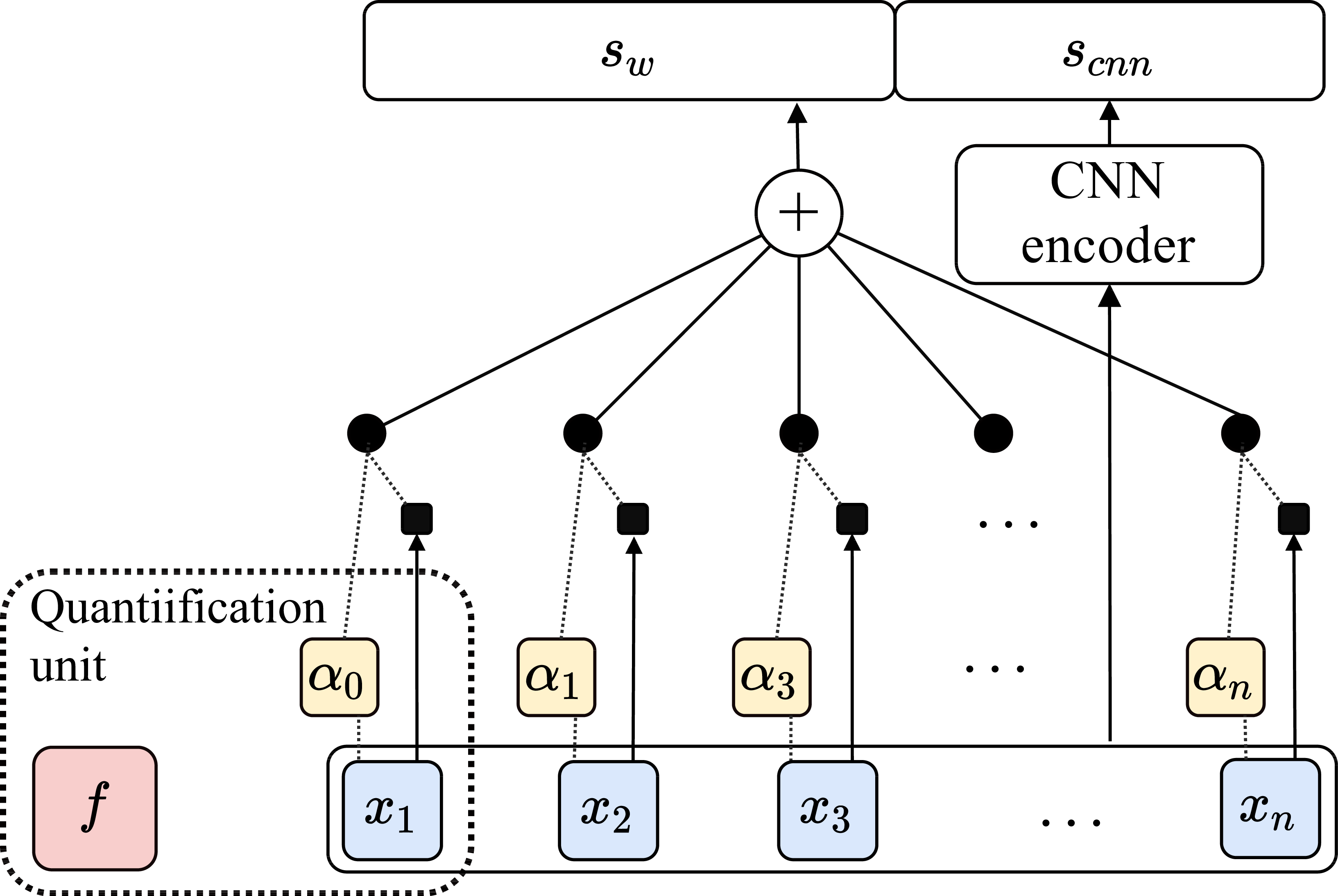}
    \caption{Feature-attentive context encoder architecture, based on \attCnn{} model~\cite{huang2016attention}.}
    \label{fig:mlp-attention-encoder}
\end{figure}

Figure~\ref{fig:mlp-attention-encoder} illustrates a feature-based encoder
~\cite{huang2016attention}.
In formulas~\ref{eq:mlp_concat}--\ref{eq:mlp_emb}, we
describe the quantification process  of a context embedding $\embSet{}$ with respect to a particular
feature $\embFeature{}~\in~\featureSet{}$.
Given an $i$'th embedded term $\embWord{i}$,
we concatenate its representation with $\embFeature{}$:

\begin{equation}
    \hidden{i} = \left[\embWord{i}, \embFeature{} \right]
    \label{eq:mlp_concat}
\end{equation}

The quantification of the relevance of $\embWord{i}$ with respect to $\embFeature{}$
is denoted as $u_{i}~\in~\mathbb{R}$ and calculated as follows:
\begin{equation}
    u_{i} = W_a\left(\tanh(W_{we} \cdot h_{i} + b_{we})\right) + b_a \hspace{0.5cm}
    \label{eq:mlp_weight}
\end{equation}

In Formula~\ref{eq:mlp_weight}, $W_{we}\in \mathbb{R}^{2 \cdot m \times \mlpHidden}$ and $W_a \in \mathbb{R}^{\mlpHidden{}}$ 
correspond to the weight and attention matrices respectively, and 
$\mlpHidden{}$ corresponds to the size of the hidden representation in the weight matrix.
To deal with normalized weights within a context, we transform quantified values $u_{i}$ 
into probabilities $\alpha_{i}$ by Formula~\ref{eq:softmax} as follows: $\alpha~=~\sigma(u, \ctxSize{}$).
We utilize Formula~\ref{eq:mlp_emb} to obtain attention-based context embedding $\featEmb{}$
of a context with respect to feature $\embFeature{}$:
\begin{equation}
    \featEmb{} = \sum_{i=1}^{\ctxSize{}} \embWord{i} \cdot \alpha_{i}
    \hspace{1cm}
    \featEmb{} \in \mathbb{R}^{\embSize{}}
    \label{eq:mlp_emb}
\end{equation}

Applying Formula~\ref{eq:mlp_emb} towards each feature $\embFeature{j} \in \featureSet{}, \hspace{0.1cm} j \in \overline{1..\featuresCount{}}$
results in vector $\{\featEmb{}_j\}_{j=1}^{\featuresCount{}}$.
We use \textit{average-pooling} to transform the latter sequence into
single averaged vector $\featureBasedVector{} \in \mathbb{R}^\embSize{}$.

We also utilize a <<CNN encoder>> block (Figure~\ref{fig:mlp-attention-encoder})
in order to compose the context representation $\cnnBasedVector{}$.
The resulting context embedding vector 
$\sentenceEmbedding{}$
is a concatenation of $\featureBasedVector{}$ and $\cnnBasedVector{}$:
\begin{equation}
    \sentenceEmbedding{}=[\featureBasedVector{}, \cnnBasedVector{}]
\end{equation}

\newcommand{\convFilter}[1]{\omega_{#1}}
\newcommand{\convLayersCount}{t}
\newcommand{\convFiltersSet}{\{\convFilter{1}, \ldots, \convFilter{\convLayersCount{}}\}}
\newcommand{\filterSize}{l}
\newcommand{\conv}[1]{c_{#1}}
\newcommand{\layerIndex}{i}
\newcommand{\embIndex}{j}
\newcommand{\embIndexLeft}{a}
\newcommand{\embIndexRight}{b}
\newcommand{\embRange}[2]{\embWord{#1:#2}}
\newcommand{\convLayer}[1]{{\bf c}_{#1}}
\newcommand{\convLayerEq}[2]{\{c_{#1,1}, \ldots, c_{#1,#2}\}}  
\newcommand{\convMatrixEq}{\{ \convLayer{1}, \convLayer{2}, \ldots, \convLayer{\convLayersCount} \}}
\newcommand{\convMatrixVar}{C}
\newcommand{\ppooling}[1]{{\bf p}_{#1}}
\newcommand{\pmPoolingPartIndex}{q}
\newcommand{\pmPoolingEq}{\{p_{\layerIndex{}, 1}, p_{\layerIndex{}, 2}, p_{\layerIndex{}, 3}\}}
\newcommand{\embIndexRange}{\embIndex{} \in \overline{1..\ctxSize{}}}
\newcommand{\layersRange}{\layerIndex{} \in \overline{1..\convLayersCount{}}}

\begin{figure}[t]
    \begin{subfigure}{.47\linewidth}
      \centering
      \includegraphics[width=\linewidth]{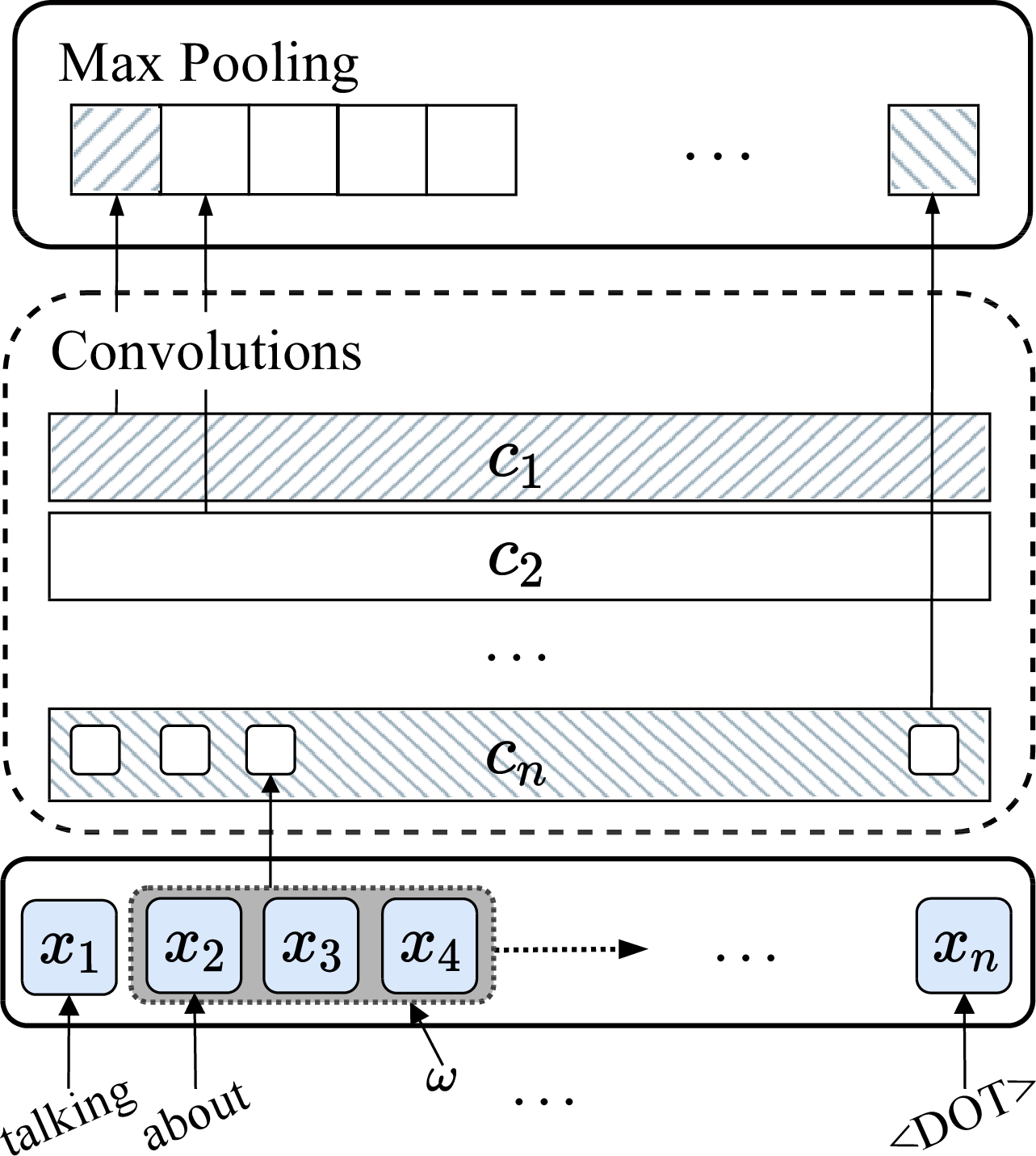}
      \caption{CNN}
      \label{fig:encoder-cnn}
    \end{subfigure} \hspace{0.3cm}
    \begin{subfigure}{.47\linewidth}
      \centering
      \includegraphics[width=\linewidth]{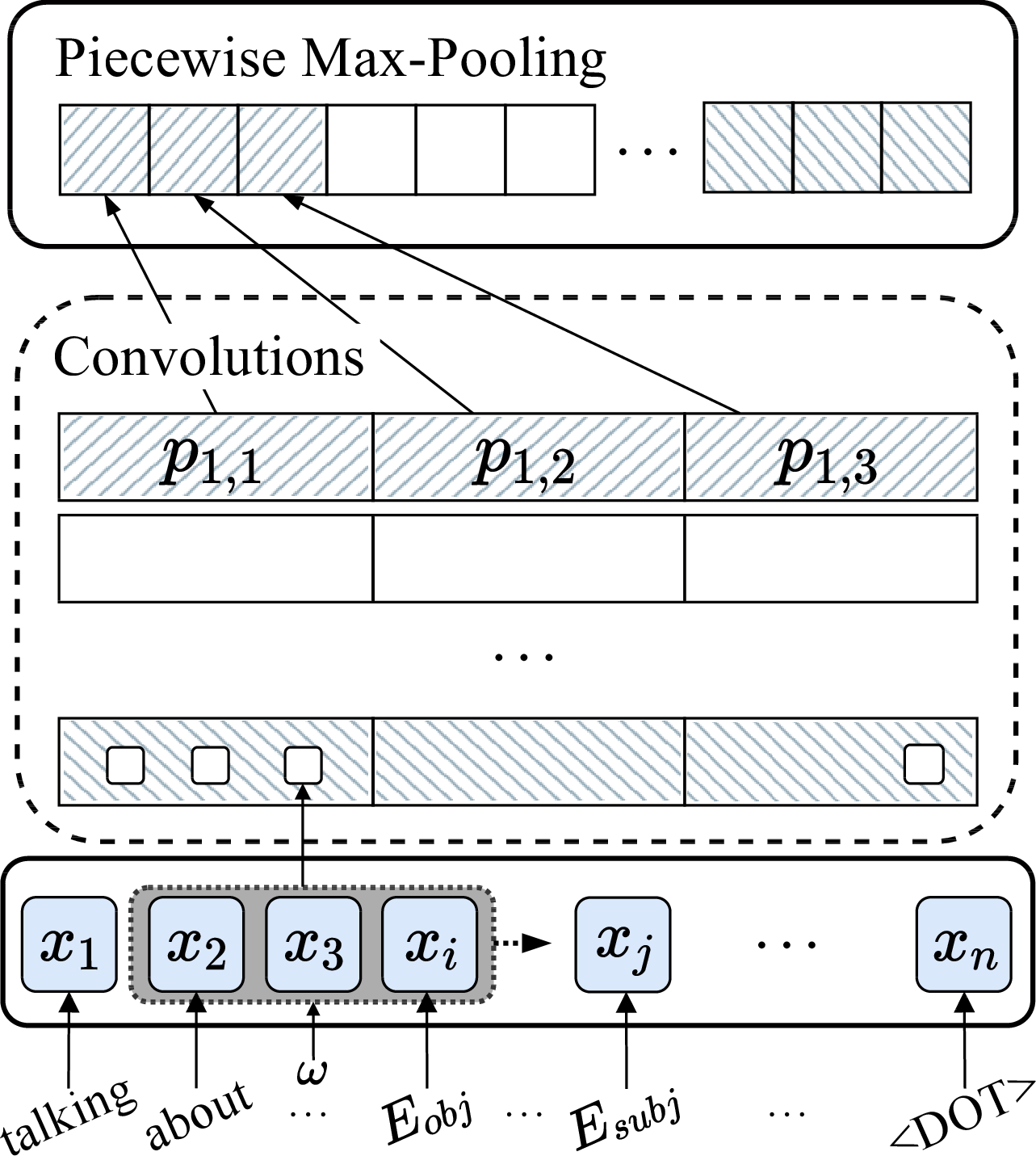}
      \caption{PCNN~\cite{zeng2015distant}}
      \label{fig:encoder-pcnn}
    \end{subfigure}
    \caption{Comparison of CNN-based context encoders; $\convFilter{}$ corresponds to convolutional filter window, size of 3.}
    \label{fig:mlp_encoders}
\end{figure}

Structurally, a convolutional neural network based encoder is a sequence of the following transformations:
convolutions and pooling.
Figure~\ref{fig:mlp_encoders} provides a detailed comparison between classic neural network
(\cnn{}, Figure~\ref{fig:encoder-cnn}),
and piecewise convolutional neural network (\pcnn{}, Figure~\ref{fig:encoder-pcnn}).

Starting with the convolution operation, which remains equal across all the encoders of Figure~\ref{fig:mlp_encoders},
let $\embRange{\embIndexLeft{}}{\embIndexRight{}}$ is as consequent vectors concatenation
from $\embIndexLeft{}$'th till $\embIndexRight{}$'th positions.
An application of ${\bf \omega} \in \mathbb{R}^d, \hspace{0.1cm} (d = \filterSize{} \cdot \embSize{})$ towards the
concatenation $\embRange{\embIndexLeft{}}{\embIndexRight{}}$ is a sequence {\it convolution} by filter $\convFilter{}$,
where $\filterSize{}$ is a filter
window size, and $\embSize{}$ corresponds to embedding vector size.
For convolving calculation $\conv{\embIndex{}}$ ($\embIndexRange{}$), we apply scalar multiplication as follows:
\begin{equation}
    \conv{\embIndex{}} = \convFilter{} \cdot \embRange{\embIndex{}-\filterSize{}+1}{\embIndex{}}
    \label{eq:convolution}
\end{equation}

To get multiple feature combinations, a set of different filters $W~=~\convFiltersSet{}$
has been applied towards $\embSet{}$.
This leads to a modified version of Formula~\ref{eq:convolution} by introduced layer index $\layerIndex{}$:
\begin{equation}
    \conv{\layerIndex{},\embIndex{}} = \convFilter{\layerIndex{}} \cdot \embRange{\embIndex{}-\filterSize{}+1}{\embIndex{}}
    \label{eq:conv2}
\end{equation}

Denoting $\convLayer{i} = \convLayerEq{i}{n}$ in Formula~\ref{eq:conv2} we reduce
the latter by index $\embIndexRange{}$ and compose a matrix $\convMatrixVar{}~=~\convMatrixEq{}$ which
represents convolution matrix with shape $\convMatrixVar{}~\in~\mathbb{R}^{\ctxSize{} \times \convLayersCount{}}$.

Max-pooling is an operation that reduces values by keeping maximum. 
In original CNN architecture (Figure~\ref{fig:encoder-cnn}), max pooling
applies separately per each convolution layers $\convLayer{\layerIndex{}}$,
which results in $\ppooling{}~\in~\mathbb{R}^{\convLayersCount{}}$.
It reduces convolved information quite rapidly which is not
appropriate for attitude classification task.
To keep context features that are inside and outside of the attitude entities,
authors~\cite{zeng2015distant} perform {\it piecewise max-pooling} (Figure~\ref{fig:encoder-pcnn}).
Given attitude entities as borders, we divide each $\convLayer{\layerIndex{}}$ into inner, left and
right segments
$\{\convLayer{\layerIndex{}, 1}, \convLayer{\layerIndex{},2}, \convLayer{\layerIndex{},3}\}$.
Then max-pooling applies per each segment separately:
\begin{equation}
    p_{\layerIndex{},\pmPoolingPartIndex{}} = \max(\convLayer{\layerIndex{}, \pmPoolingPartIndex{}}),
    \hspace{0.2cm}
    \layersRange{},
    \hspace{0.2cm}
    \pmPoolingPartIndex{} \in \{1, 2, 3\}
\end{equation}

Thus, for each $\convLayer{\layerIndex{}}$ we have a set $\ppooling{\layerIndex{}}~=~\pmPoolingEq{}$.
Concatenation of these sets for each layer $\layerIndex{}$ results in $\ppooling{} \in \mathbb{R}^{3\convLayersCount{}}$
and that is a result of piecewise max-pooling operation.

\section{Self Attentive Context Encoders}
\label{sec:self-based}

In section~\ref{sec:mlp-based} the application of attention in context embedding fully relies on the sequence of predefined features.
The quantification of context terms is performed towards each feature.
In turn, the \textit{self-attentive} approach assumes to quantify a context with respect to an
abstract parameter.
Unlike quantification methods in feature-attentive embedding models,
here the latter is replaced with a hidden state ($w$)
which modified during the training process.

\newcommand{\selfHidden}{\textbf{h}}

\newcommand{\selfHiddenVector}{w}

\newcommand{\selfAttHidden}{H_a}

\begin{figure}[!t]
    \captionsetup[subfigure]{position=b}
    \centering
    \includegraphics[width=0.97\linewidth]{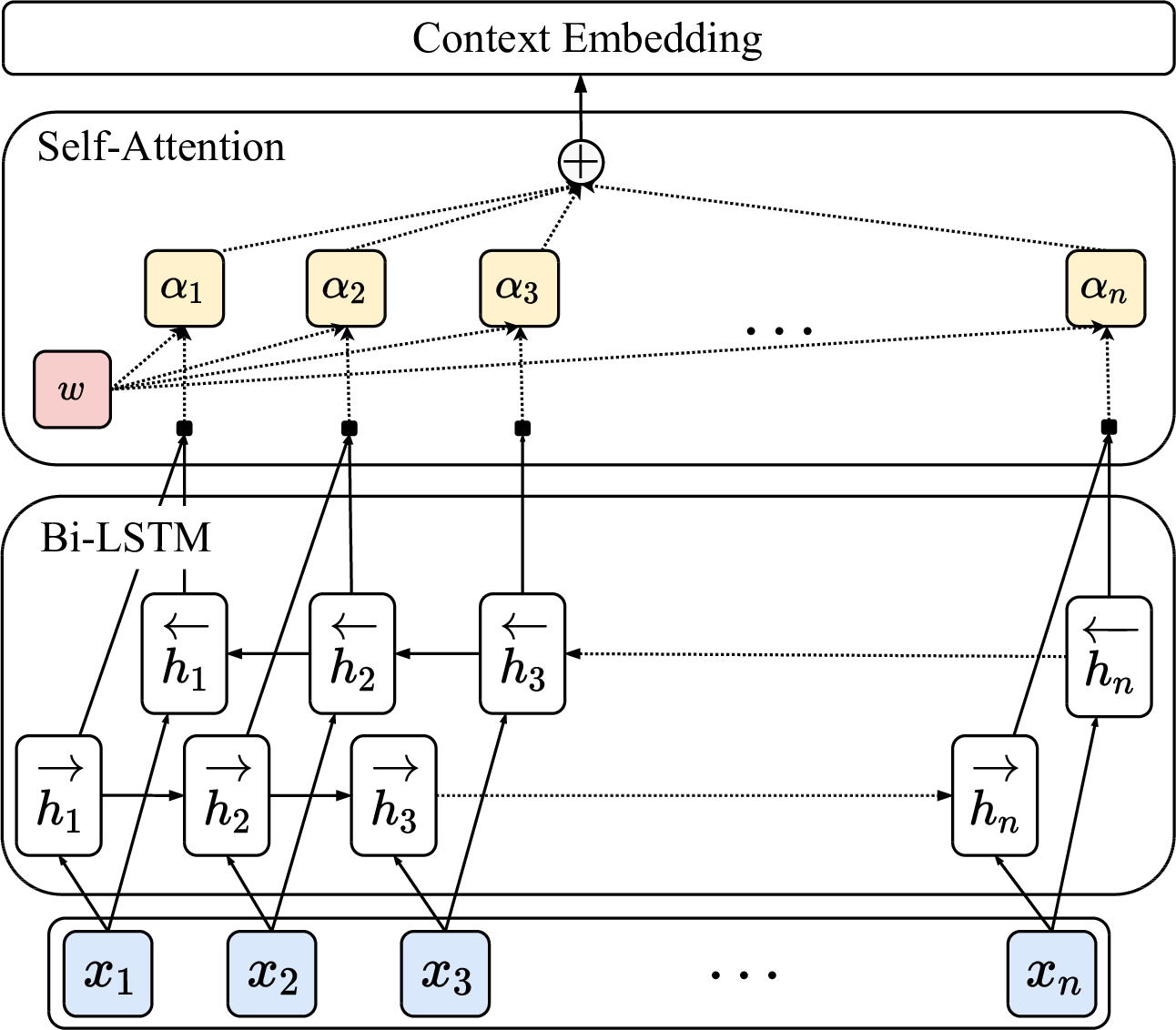}
    \caption{Self-attentive context encoder architecture,
    with self-attention module of \bilstmPZhou{} model~\cite{zhou2016attention} over bi-directional \lstm{} encoder.}
    \label{fig:self-based}
\end{figure}

To learn the hidden term semantics for each input,
we utilize the \lstm{}~\cite{hochreiter1997long} recurrent neural network architecture, which addresses
learning long-term dependencies by avoiding gradient vanishing and expansion problems.
The calculation $\hidden{t}$ of $t$'th embedded term $\embWord{t}$ is based
on prior state $\hidden{t-1}$, where the latter acts as a parameter of
auxiliary functions~\cite{hochreiter1997long}.
Figure~\ref{fig:self-based} illustrates the attention-based sentence encoder architecture,
builded on top of the \bilstm{} -- is a bi-directional \lstm{} to obtain a pair of sequences
$\overrightarrow{\hidden{}}$
and
$\overleftarrow{\hidden{}}$ \hspace{0.5mm}
($\overrightarrow{\hidden{i}}, \overleftarrow{\hidden{i}} \in \mathbb{R}^{\selfHidden{}}$).
The resulting context representation $H=[\hidden{1},~\ldots,~\hidden{\ctxSize{}}]$ is
composed as the concatenation of bi-directional sequences elementwise:
$\hidden{i} = \overrightarrow{\hidden{i}} + \overleftarrow{\hidden{i}}, \hspace{0.1cm}
i \in \overline{1..\ctxSize{}}$.
The quantification of hidden term representation $\hidden{i} \in \mathbb{R}^{2 \cdot \selfHidden{}}$
with respect to $w \in \mathbb{R}^{2 \cdot \selfHidden{}}$ is
described in formulas~\ref{eq:self-q1}-\ref{eq:self-q2}.
\begin{equation}
    m_i = \tanh(\hidden{i})
    \label{eq:self-q1}
\end{equation}
\begin{equation}
    u_i = m_i^T \cdot w
    \label{eq:self-q2}
\end{equation}

In order to deal with normalized weights, we transoform quantified values $u_i$
into $\alpha_i$ as follows: $\alpha=\sigma(u, \ctxSize{})$ (Formula~\ref{eq:softmax}).
The resulting context embedding vector
$\sentenceEmbedding{}$
is an activated weighted sum of each parameter of  context hidden states:
\begin{equation}
    \sentenceEmbedding{} = tanh(H \cdot \alpha)
    \hspace{1cm}
    \sentenceEmbedding{} \in \mathbb{R}^{2 \cdot \selfHidden{}}
\end{equation}
\section{Model Details}
\newcommand{\fmtest}{\fm{\test{}}}

\begin{table*}[!htp]
\centering

\begin{tabular}{lc|c|ccc|c||c|ccc|c}

                         &            & \multicolumn{5}{c||}{\textsc{\twoScale{}}} & \multicolumn{5}{c}{\threeScale{}} \\ \hline
&&&&&&&&&&&\\ [-0.95em]
Model                    & DS         & $\fm{avg}$   & $\fmCv{1}$    & $\fmCv{2}$    & $\fmCv{3}$    & $\fm{\test{}}$ & $\fm{avg}$   & $\fmCv{1}$    & $\fmCv{2}$    & $\fmCv{3}$    & $\fm{\test{}}$ \\
\hline
\bilstmPZhou{}           & \useDS{} &        \textbf{0.667} & \textbf{0.71} &          0.62 &          0.67 & 0.68  &   \textbf{0.332} &      \textbf{0.36} & \textbf{0.33} &         0.31  &  0.38          \\
\bilstm{}                & \useDS{} &        0.653 &          0.70 &          0.60 &          0.66 & \textbf{0.70} &        0.312 &          0.34 &          0.31 &          0.29 &  0.39          \\
\bilstmPZhou{}           &          &        0.640 &          0.69 &          0.60 &          0.64 & 0.68  &        0.314 &          0.35 &          0.27 & \textbf{0.32} &  0.32          \\
\bilstm{}                &          &        0.632 &          0.66 &          0.63 &          0.61 & 0.67  &        0.286 &          0.32 &          0.26 &          0.28 &  0.34          \\
\hline
\pcnnEnds{}              & \useDS{} &        0.644 &          0.67 &          0.61 &          0.65 & 0.66  &        0.312 &          0.33 &          0.30 &          0.31 &  \textbf{0.41} \\
\pcnn{}                  & \useDS{} &        0.599 &          0.70 &          0.53 &          0.57 & 0.63  &        0.315 &          0.33 &          0.30 &          0.31 &  0.40          \\
\pcnnEnds{}              &          &        0.617 &          0.64 &          0.56 &          0.65 & 0.67  &        0.297 &          0.32 &          0.29 &          0.28 &  0.35          \\
\pcnn{}                  &          &        0.608 &          0.62 &          0.58 &          0.63 & 0.66  &        0.285 &          0.29 &          0.27 &          0.30 &  0.32          \\
\hline
\cnnEnds{}               & \useDS{} &        0.631 &          0.64 &          0.64 &          0.62 & 0.66  &        0.316 &          0.35 &          0.29 &          0.30 &  \textbf{0.41} \\
\cnn{}                   & \useDS{} &        0.625 &          0.62 &          0.63 &          0.63 & 0.68  &        0.305 &          0.31 &          0.30 &          0.31 &  0.40          \\
\cnnEnds{}               &          &        0.636 &          0.66 &          0.64 &          0.61 & 0.62  &        0.270 &          0.33 &          0.23 &          0.25 &  0.30          \\
\cnn{}                   &          &        0.553 &           0.60 &          0.56 &          0.51 & 0.59  &        0.274 &          0.30 &          0.26 &          0.26 &  0.31          \\
\hline
\end{tabular}

\caption{Experiment (\twoScale{} and \threeScale{}) context classification results by $\fm{}$
measure over \rusentrel{}~collection;
all the models are separated into the following groups (from top to bottom): \bilstm{}, \pcnn{}, \cnn{};
models that employ~\ruattitudes{} in the training process (\dsm{}~mode) are labeled with <<\useDS{}>>;
columns related to result evaluation in each experiment (from left to right):
(1)~average value in CV-3 experiment ($\fm{avg}$) with results on each split
($\fmCv{i}, \hspace{0.1cm} i \in \overline{1..3}$),
(2)~results on \train{}/\test{} separation ($\fmtest{}$).}
\label{tab:results_combined}

\end{table*}

\label{sec:model_details}
\label{sec:input-embedding-details}

\newcommand{\featureSize}{5}
\newcommand{\embeddingWindowSize}{20}
\newcommand{\embeddingVectorSize}{1000}
\newcommand{\wordEmbedding}{$M_{word}$}

\newcommand{\featurePOS}{$v_\textsc{pos}$}
\newcommand{\featureDistObj}{$v_{\textsc{d}\mhyphen obj}$}
\newcommand{\featureDistSubj}{$v_{\textsc{d}\mhyphen subj}$}
\newcommand{\featureSynDistObj}{$v_{\textsc{sd}\mhyphen obj}$}
\newcommand{\featureSynDistSubj}{$v_{\textsc{sd}\mhyphen subj}$}
\newcommand{\featurePolarity}{$v_{A0\to A1}$}

We provide embedding details of context term groups described in Section~\ref{sec:model}.
For \wordsGroup{} and \framesGroup{}, we look up for vectors in precomputed and publicly available model\footnote{
\url{http://rusvectores.org/static/models/rusvectores2/news_mystem_skipgram_1000_20_2015.bin.gz}}
\wordEmbedding{} based on news articles with 
window size of $\embeddingWindowSize{}$,
and vector size of $\embeddingVectorSize{}$.
Each term that is not presented in model we treat as a sequence of \textit{parts} ($n$-grams)
and look up for related vectors in~\wordEmbedding{} to complete an averaged vector.
For a particular part, we start with trigrams ($n=3$) and decrease $n$ until the related $n$-gram is found.
For masked entities (\maskE{}, \maskEObj{}, \maskESubj{}) and ~\tokenGroup{}, each element embedded with a vector of size 1000; every vector is randomly initialized from a Gaussian distribution~\cite{glorot2010understanding}.

Each context term has been additionally expanded with the following parameters:
\begin{itemize}
    \item Distance embedding~\cite{rusnachenko2018neural} (\featureDistObj{}, \featureDistSubj{})
    -- is vectorized distance in terms from attitude participants of entry pair (\maskEObj{} and \maskESubj{} respectively) 
    to a given term;
    \item Closest to synonym distance embedding (\featureSynDistObj{}, \featureSynDistSubj{})
    is a vectorized absolute distance in terms from
    a given term towards the nearest entity, synonymous to \maskEObj{} and \maskESubj{}; 
    \item Part-of-speech embedding (\featurePOS{})
    is a vectorized tag  for \wordsGroup{} (for terms of other groups considering <<unknown>> tag);
    \item \polarity{} polarity embedding (\featurePolarity{}) is
    a vectorized <<positive>> or <<negative>> value for frame entries whose description in \rusentiframes{} provides the corresponding polarity
    (otherwise considering <<neutral>> value);
    polarity is inverted when an entry has \begin{otherlanguage*}{russian}"не"\end{otherlanguage*} (not) preposition.
\end{itemize}


\subsection{Training}

\newcommand{\lossF}[1]{L_{#1}}

This process assumes hidden parameter optimization of a given model.
We utilize an algorithm described in~\cite{rusnachenko2018neural}.
The input is organized in minibatches, where each minibatch yields of $\bagsCount{}$ \textit{bags}.
Each bag has a set of $\sentencesCount{}$ pairs
$\left<\embSet{}_j, y_j\right>_{j=1}^{\sentencesCount{}}$,
where each pair is described by an input embedding $\embSet{}_j$
with the related label $y_j\in\mathbb{R}^\classesCount{}$.
The training process is iterative, and each iteration includes the following steps
in order to calculate vector $cost$ and perform hidden states update.

The first step assumes a minibatch composing, which is consist of $\bagsCount{}$ bags of size $\sentencesCount{}$.
Then we perform a forward propagation through the network which results in
a vector (size of $q = \bagsCount{} \cdot \sentencesCount{}$) 
of outputs $o_k\in\mathbb{R}^\labelsCount{}$.
In the third step we calculate \textit{cross entropy loss} for an output vector as follows:
\begin{equation}
    \lossF{k} = \sum\limits_{j=1}^\labelsCount{} \log p(y_i|o_{k,j}; \theta), \hspace{0.1cm} k \in \overline{1..q}
\end{equation}
In the final step we compose a $cost$ vector, where $i$'th component $cost_i$ ($i=\overline{1..\bagsCount{}}$) corresponds
to the maximal cross entropy loss within a related $i$'th bag:
\begin{equation}
    cost_i = \max\left[\lossF{(i-1) \cdot \sentencesCount} \hspace{0.1cm} .. \hspace{0.1cm} \lossF{i\cdot \sentencesCount}\right)
    \label{eq:cost}
\end{equation}
\subsection{Parameters settings}
\label{sec:params}

\newcommand{\bagsCountValue}{$2$}
\newcommand{\ctxPerBag}{$3$}
\newcommand{\termsPerContext}{50}
\newcommand{\pairDistValue}{10}
\newcommand{\framesPerContext}{$5$}
\newcommand{\lstmHidden}{128}

\newcommand{\cnnWindowSizeValue}{$3$}
\newcommand{\cnnFiltersCountValue}{$300$}

\newcommand{\dropoutKeepProb}{$0.8$}

The minibatch size ($\bagsCount{}$) is set to \bagsCountValue{}, 
where contexts count per bag $\sentencesCount{}$ is set to \ctxPerBag{}.
All the contexts were limited by $\ctxSize{} = \termsPerContext{}$ terms,
with the distance between attitude participants limited to $\pairDist{} = \pairDistValue{}$ terms.
For embedding parameters (Section~\ref{sec:input-embedding-details})
we use vectors with size of~$\featureSize{}$.
For \cnn{} and \pcnn{} context encoders, 
the size of convolutional window ($\omega$)
and filters count (\convCount{})
were set to \cnnWindowSizeValue{} and \cnnFiltersCountValue{} respectively. 
As for parameters related to sizes of hidden states in Sections~\ref{sec:mlp-based} and \ref{sec:self-based}:
$\mlpHidden{}=10$, 
$\ianHidden{}=\lstmHidden{}$. 
We utilize the AdaDelta optimizer with parameters $\rho=0.95$ and $\epsilon=10^{-6}$~\cite{zeiler2012adadelta}.
To prevent models from overfitting, 
we apply $dropout$ towards the output
with keep probability set to~\dropoutKeepProb{}.
For hidden state values initialization we utilize Xavier weight intializer~\cite{glorot2010understanding}.

\section{Experiments}
\label{sec:experiments}

\newcommand{\rDs}{$E_{DS}$}
\newcommand{\rDsAtt}{$E_{DSA}$}
\newcommand{\cvR}{\textsc{CV-based}}
\newcommand{\fR}{\textsc{Fixed}}

According to Section~\ref{sec:model}, we treat sentiment attitude extraction
as a classification task of different scales of output classes.
We train and evaluate all the models in the following experiments:
\begin{enumerate}
    \item \twoScale{}~\cite{rusnachenko2019distant}, in which all the models
    have to predict a sentiment label of an attitude in context.
    It is important to note that for each document we consider only those attitudes that
    might be fitted in a context;
    \item \threeScale{}~\cite{rusnachenko2018neural}, in which each model might classify a given context with an attitude
    in it as sentiment-oriented (positive/negative) or \textit{neutral}.
\end{enumerate}

It is worth to note that the evaluation process in  case of \twoScale{} experiment
assumes to treat only those pairs in comparison, which could be found within a context of the related document.

\subsection{Datasets and Evaluation formats}
\label{sec:evaluation}

The evaluation in experiments has been performed over the \rusentrel{} corpus, using the following formats:
\begin{enumerate}
    \item \cvR{} format, in which it is supposed to utilize 3-fold cross-validation (CV);
    all  folds are equal in terms of sentence count;
    \item \fR{} format, in which the predefined separation of documents onto \train{}/\test{} sets is considered\footnote{
        \url{https://miem.hse.ru/clschool/results}}.
\end{enumerate}

For evaluating models in this task,
we adopt macro-averaged F1-score (\fmeasure{}) over documents.
F1-score is considered averaging of the positive and negative classes, which are most important in attitude analysis.

\subsection{Model Comparisons and Training}
\label{sec:modelcmp}

In terms of architecture aspects, all the models differ only in sentence encoder
implementation of a single context classification model (Figure~\ref{fig:model}).
The list of the models selected for the experiments is as follows:
\begin{itemize}
    \item \textbf{\cnn{}} model with a classic convolutional neural network architecture (Figure~\ref{fig:encoder-cnn});
    \item \textbf{\pcnn{}} model, in which the encoder treats each convolution layer in parts,
    relatively to the attitude participants' positions in the context (Figure~\ref{fig:encoder-pcnn});
    \item \textbf{\cnnEnds{}}, \textbf{\pcnnEnds{}} are models with feature attentive encoders (Section~\ref{sec:mlp-based}); <<e>> corresponds to the set of attitude participants (\maskEObj{}, \maskESubj{}).
    \item \textbf{\bilstm{}} is a bi-directional \lstm{}~\cite{hochreiter1997long};
    \item \textbf{\bilstmPZhou{}} model 
    (Section~\ref{sec:self-based});
\end{itemize}

For a particular model, the training (and related evaluation) process
has been performed in the following modes:
\begin{enumerate}
    \item \dsm{}, is an application of distant supervision, which is considered as a combination of \rusentrel{} and \ruattitudes{} collections;
    \item \slm{}, is supervised learning, using \rusentrel{}.
\end{enumerate}

It is worth to clarify the details of the training set creation in \textsc{DS} mode depending on the evaluation formats (Section~\ref{sec:evaluation}):
\begin{itemize}
    \item For \cvR{}, in each split, the \ruattitudes{} collection is combined with each training block of the \rusentrel{} collection;
    \item For \fR{}, the training set represents a combination of \ruattitudes{} with the \train{} part.
\end{itemize}

We measure \fmeasure{} on the training part every \epochsToTest{} epoch.
The number of epochs was limited by \epochsCount{}.
The training process terminates when \fmeasure{} on the training part becomes greater than $0.85$.

\subsection{Result Analysis}

Table~\ref{tab:results_combined} provides the results in the  experiments for models organized (and separated)
into the following groups: \cnn{}, \pcnn{}, \bilstm{}.
To access the effectiveness of both an application of
distant supervision in the training process
(\textsc{DS} mode, marked with <<\useDS{}>> sign in Table~\ref{tab:results_combined})
and attention-based encoders (prefixed with <<\textsc{Att}>>),
we provide efficiency assessment in the following directions:
\begin{enumerate}
    \item Application of \dsm{} mode for baselines;
    \item Application of attention-based sentence encoders in \dsm{} mode.
\end{enumerate}

\newcommand{\toPer}[2]{round(#1/#2 - 1, 2)}

\newcommand{\placeholderBase}{0.01}
\newcommand{\placeholderAtt}{0.02}

\begin{table}[!t]
\begin{center}
\begin{tabular}{c|c|ccc|ccc}
    \multicolumn{2}{c|}{}    & \multicolumn{3}{c|}{\twoScale{}} & \multicolumn{3}{c}{\threeScale{}} \\ \hline
    & & & & & & &\\
    Ratio      & Parameter   & \rotatebox[origin=c]{90}{\cnn{}}  &  \rotatebox[origin=c]{90}{\pcnn{}}  & \rotatebox[origin=c]{90}{\bilstm{}} & \rotatebox[origin=c]{90}{\cnn{}}  &  \rotatebox[origin=c]{90}{\pcnn{}}  & \rotatebox[origin=c]{90}{\bilstm{}}      \\ 
    & & & & & & &\\
    \hline
    \multirow{2}{*}{\rDs{}}  & $\fm{avg}$
                             & \FPeval{\result}{\toPer{0.625}{0.553}} \result
                             & \no{} 
                             & \placeholderBase{}
                             & \FPeval{\result}{\toPer{0.305}{0.274}} \result
                             & \FPeval{\result}{\toPer{0.315}{0.285}} \result
                             & \FPeval{\result}{\toPer{0.312}{0.286}} \result
                \\

               & $\fmtest{} $ & \FPeval{\result}{\toPer{0.68}{0.59}} \result
                              & \no{} 
                              & \FPeval{\result}{\toPer{0.70}{0.67}} \result

                              & \FPeval{\result}{\toPer{0.40}{0.31}} \result
                              & \FPeval{\result}{\toPer{0.40}{0.32}} \result
                              & \FPeval{\result}{\toPer{0.39}{0.34}} \result
                \\ \hline

    \multirow{2}{*}{\rDsAtt{}}& $\fm{avg}$   & \FPeval{\result}{\toPer{0.631}{0.625}} \result
                              & \FPeval{\result}{\toPer{0.644}{0.599}}  \result
                              & \placeholderAtt{}

                              & \FPeval{\result}{\toPer{0.316}{0.305}} \result
                              & \no{} 
                              & \FPeval{\result}{\toPer{0.332}{0.312}} \result
                 \\

               & $\fmtest{} $ & \no{} 
                              & \FPeval{\result}{\toPer{0.66}{0.63}} \result
                              & \no{} 

                              & \FPeval{\result}{\toPer{0.41}{0.40}} \result
                              & \FPeval{\result}{\toPer{0.41}{0.40}} \result
                              & \no{} 
                \\ \hline
\end{tabular}
\end{center}
    \caption{
        Calculated \rDs{} and \rDsAtt{} ratios in each experiment
        for \cvR{} ($\fm{avg}$) and \fR{} ($\fmtest{}$) evaluation formats;
        values below zero displayed as <<\no>>}
    \label{tab:stat}
\end{table}

To accomplish the comparison in a particular experiment, for each model we calculate the corresponding ratios by $\fm{avg}$ and $\fmtest{}$:
\begin{itemize}
    \item \rDs{} -- is the effectiveness of baseline models trained in \dsm{} mode
    over a related baseline that trained in \slm{} mode;
    \item \rDsAtt{} -- is the effectiveness of models trained in \dsm{} mode
    with attention-based sentence encoder (prefixed with~\textsc{Att})
    over related baseline version.
\end{itemize}



Table~\ref{tab:stat} provides calculated ratios for the \twoScale{} and \threeScale{} experiments.
The ratio calculation ($r$) for a result $A$ over a result $B$ performed as follows: $r = A/B - 1$.

Analyzing results in the \twoScale{} experiment by \rDs{} in Table~\ref{tab:stat}, model
\cnnEnds{} shows a significant increase in 13\% and 15\% in case of \cvR{} and \fR{} evaluation formats respectively.
An application of attention-based encoders does not illustrate an increase in result model quality, only 1\%
for \cnnEnds{} and 5-8\% for \pcnnEnds{}.
The highest result is obtained by the \bilstmPZhou{} model with a 4\% increase by \rDs{}.

\newcommand{\dsAttImproveRangeCVBLSTMPZhou}{6\%}

As for   the \threeScale{} experiment, it is also possible to investigate a significant increase by \rDs{}
with 10\% in the \cvR{} evaluation  mode and 15-29\% on the \test{} part (\fR{} evaluation format).
Utilizing attentive encoders in the models that employ \ruattitudes{} in training provides 3\% results improvement
according to \rDsAtt{} ratio.
The highest increase by \rDsAtt{} is achieved by \bilstmPZhou{} model with \dsAttImproveRangeCVBLSTMPZhou{} when the model is
evaluated in the \cvR{} format.

\newcommand{\modelA}{(1)}
\newcommand{\modelB}{(2)}
\newcommand{\modelC}{(3)}

\newcommand{\modelAName}{\pcnnEf{}}
\newcommand{\modelCName}{\bilstmPZhou{}}

\newcommand{\nounsGroup}{\textsc{nouns}}
\newcommand{\verbsGroup}{\textsc{verbs}}

\newcommand{\sDm}[1]{\rho_{\textsc{s}}^{#1}}
\newcommand{\nDm}[1]{\rho_{\textsc{n}}^{#1}}
\newcommand{\sD}[1]{$\sDm{#1}$}
\newcommand{\nD}[1]{$\nDm{#1}$}

\section{Analysis of Attention Weights}

According to Section~\ref{sec:ruattitudes}, one of the assumptions behind the distant supervision application for
\ruattitudes{} collection developing is that the attitude
might be conveyed by a frame of a certain sentiment polarity.
For models of the \threeScale{} experiment with attention-based encoders (\cnnEnds{}, \pcnnEnds{}, \bilstmPZhou{}),
in this section, we analyze
how contexts with sentiment and neutral attitudes affect on weight distribution in dependence on the term type.

\begin{figure*}[!t]
    \centering
    \begin{tabular}{|cccc|}
        \hline
        \multicolumn{4}{|c|}{\bilstmPZhou{} (DS)} \\ \hline
        \framesGroup{} & \nounsGroup{} & \prepGroup{} & \sentGroup{} \\
        \includegraphics[width=.18\linewidth]{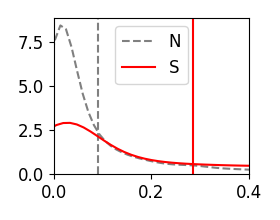} &
        \includegraphics[width=.18\linewidth]{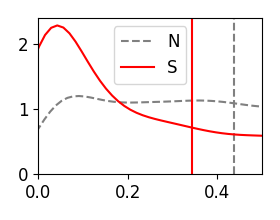} &
        \includegraphics[width=.18\linewidth]{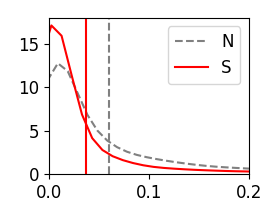} &
        \includegraphics[width=.18\linewidth]{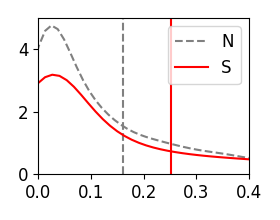} \\ [-0.6em] \hline
        \multicolumn{4}{|c|}{\bilstmPZhou{} (SL)}\\ \hline
        \framesGroup{} & \nounsGroup{} & \prepGroup{} & \sentGroup{} \\
        \includegraphics[width=.18\linewidth]{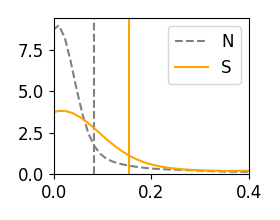} &
        \includegraphics[width=.18\linewidth]{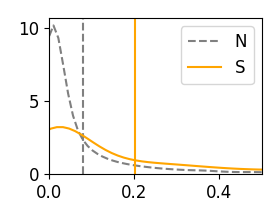} &
        \includegraphics[width=.18\linewidth]{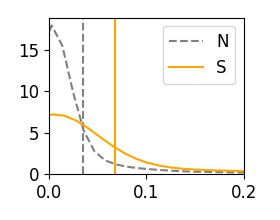} &
        \includegraphics[width=.18\linewidth]{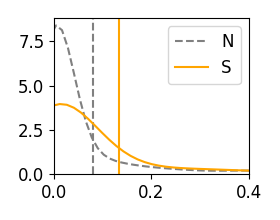} \\ \hline

        \multicolumn{4}{c}{}\\
        \hline
        \multicolumn{4}{|c|}{\cnnEnds{} (DS)}\\ \hline
        \framesGroup{} & \nounsGroup{} & \prepGroup{} & \sentGroup{} \\
        \includegraphics[width=.18\linewidth]{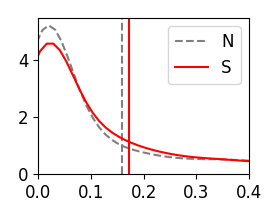} &
        \includegraphics[width=.18\linewidth]{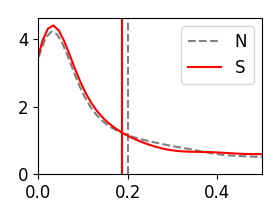} &
        \includegraphics[width=.18\linewidth]{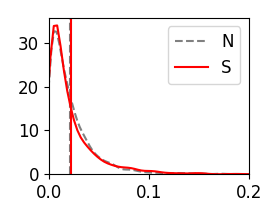} &
        \includegraphics[width=.18\linewidth]{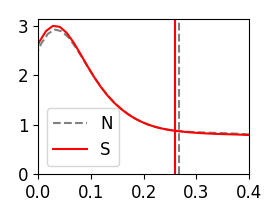} \\ [-0.6em] \hline
        \multicolumn{4}{|c|}{\cnnEnds{} (SL)}\\ \hline
        \framesGroup{} & \nounsGroup{} & \prepGroup{} & \sentGroup{} \\
        \includegraphics[width=.18\linewidth]{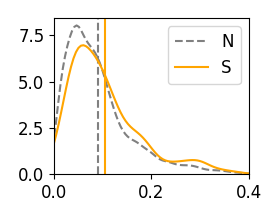} &
        \includegraphics[width=.18\linewidth]{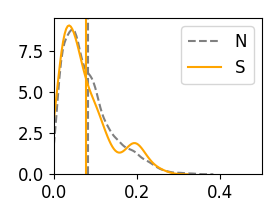} &
        \includegraphics[width=.18\linewidth]{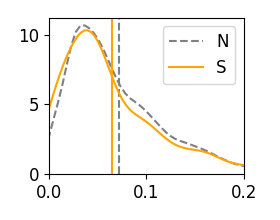} &
        \includegraphics[width=.18\linewidth]{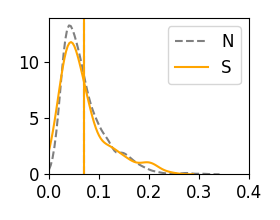} \\ \hline
    \end{tabular}
    \caption{Kernel density estimations (KDE) of context-level weight distributions
    across
    \textit{neutral} (\texttt{N})
    and
    \textit{sentiment} (\texttt{S})
    context sets for models
    \bilstmPZhou{} and
    \cnnEnds{}
    trained in different modes: distant supervision application (DS), and supervised learning only (SL);
    the probability range (x-axis) scale depends on the group of terms: $[0, 0.4]$ (\framesGroup{}, \sentGroup{}),
    $[0, 0.5]$ (\nounsGroup{}), and $[0, 0.2]$ (\prepGroup{});
    vertical lines indicate expected values of corresponding distributions.}
    \label{fig:dist-analysis}
\end{figure*}

\newcommand{\dks}[1]{$D_{#1}$}

\begin{table}[!htp]
    \begin{tabular}{lcccccc}
    \hline
    Model                   & DS         &  \dks{F}          & \dks{N}           &  \dks{P}       &  \dks{S}          &  \dks{V} \\ \hline
    \bilstmPZhou{}          & \useDS{}   &  \underline{0.29} &  \underline{0.23} &  \textbf{0.26} &  0.14             &  \textbf{0.17} \\
    \bilstmPZhou{}          &            &  0.13             &  0.22             &  0.08          &  0.11             &  0.07 \\ \hline
    \cnnEnds{}              & \useDS{}   &  0.05             &  0.03             &  0.05          &  0.03             &  0.03 \\
    \cnnEnds{}              &            &  0.09             &  0.07             &  0.09          &  0.07             &  0.07 \\ \hline
    \pcnnEnds{}             & \useDS{}   &  0.10             &  0.03             &  0.04          &  0.04             &  0.06 \\
    \pcnnEnds{}             &            &  0.09             &  0.17             &  0.15          &  0.08             &  0.06 \\ \hline
    \end{tabular}
    \caption{Calculated statistics (\dks{*}) from Kolmogorov-Smirnov test by following term groups:
    \framesGroup{} (F),
    \nounsGroup{} (N),
    \prepGroup{} (P),
    \sentGroup{} (S), and
    \verbsGroup{} (V);
    highest and second highest values per each category are bolded and underlined respectively.}
    \label{tab:ks_stat}
\end{table}

\begin{table}[!htp]
    \begin{tabular}{lrrrrrrr}
        \hline
        Model             & DS         & \multicolumn{1}{c}{$\dd{F}$} & \multicolumn{1}{c}{$\dd{N}$}
                                       & \multicolumn{1}{c}{$\dd{P}$} & \multicolumn{1}{c}{$\dd{S}$}
                                       & \multicolumn{1}{c}{$\dd{V}$} \\
        \hline
        \bilstmPZhou{}    & \useDS{}   &  \textbf{0.20}   & -0.09  & -0.02  &  \textbf{0.09}  &   \no{} \\
        \bilstmPZhou{}    &            &  0.07   &  \textbf{0.12}  &  0.03  &  0.05  &   0.03 \\ \hline
        \cnnEnds{}        & \useDS{}   &  \no{}  &  \no{} &  \no{} &  \no{} &   \no{} \\
        \cnnEnds{}        &            &  \no{}  &  \no{} &  \no{} &  \no{} &   \no{} \\ \hline
        \pcnnEnds{}       & \useDS{}   &  0.06   &  \no{} &  \no{} &  \no{} &   \no{} \\
        \pcnnEnds{}       &            &  \no{}  & -0.02  &  \no{} &  \no{} &   \no{} \\ \hline
    \end{tabular}
    \caption{The difference in estimated values of \sD{} and \nD{} ($\dd{\Any{}}$) by following term groups:
    \framesGroup{} (F),
    \nounsGroup{} (N),
    \prepGroup{} (P),
    \sentGroup{} (S), and
    \verbsGroup{} (V);
    absolute max values by each term group are bolded;
    absolute values less or equal $0.1$ displayed as~<<\no>>}
    \label{tab:mean_stat}
\end{table}

\begin{table*}[t]
    \begin{tabularx}{\textwidth}{l}
        \hline
        \multicolumn{1}{c}{\modelCName{} (SL) (Original)} \\ \hline
        \includegraphics[width=0.70\textwidth]{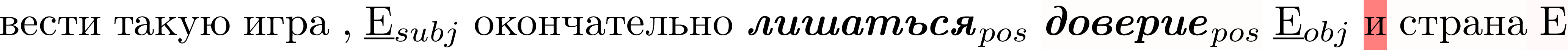} \\ [-0.7em]
        \multicolumn{1}{c}{\textbf{\ldots}} \\  [-0.1em]
        \includegraphics[width=0.99\textwidth]{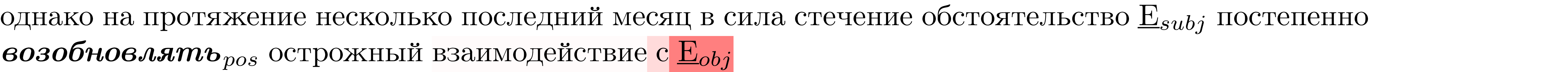} \\ [-0.7em]
        \multicolumn{1}{c}{\textbf{\ldots}} \\  [-0.1em]
        \includegraphics[width=0.99\textwidth]{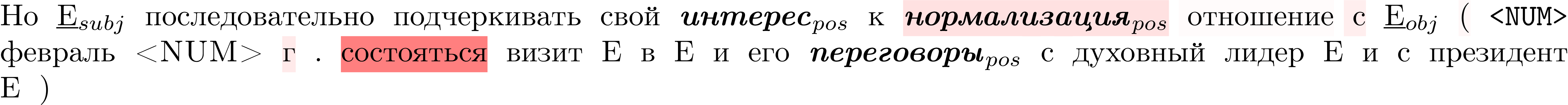} \\ \hline

        \multicolumn{1}{c}{\modelCName{} (SL)} \\ \hline
        \includegraphics[width=0.70\textwidth]{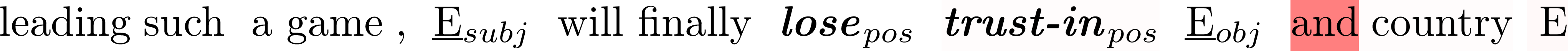} \\ [-0.7em]
        \multicolumn{1}{c}{\textbf{\ldots}} \\  [-0.1em]
        \includegraphics[width=0.99\textwidth]{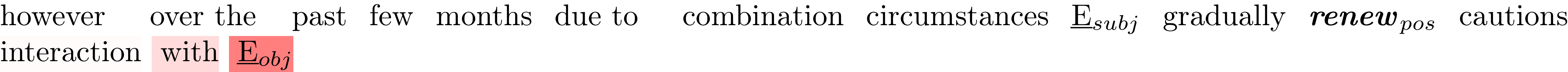} \\ [-0.7em]
        \multicolumn{1}{c}{\textbf{\ldots}} \\  [-0.1em]
        \includegraphics[width=0.99\textwidth]{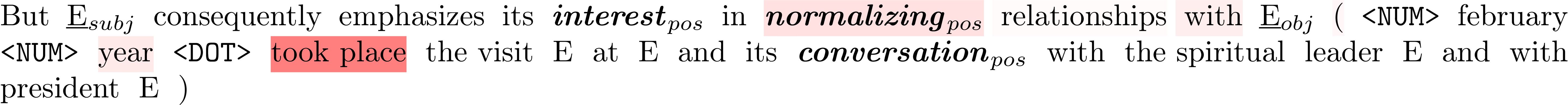} \\ \hline

        \multicolumn{1}{c}{\modelCName{} (DS)} \\ \hline
        \includegraphics[width=0.70\textwidth]{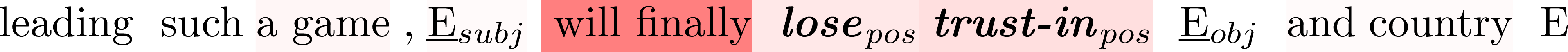} \\ [-0.7em]
        \multicolumn{1}{c}{\textbf{\ldots}} \\  [-0.1em]
        \includegraphics[width=0.99\textwidth]{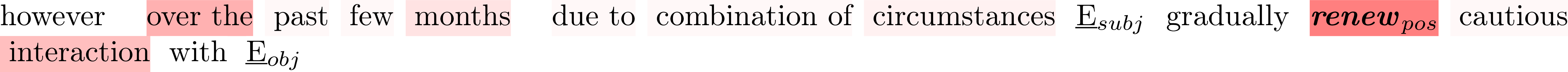} \\ [-0.7em]
        \multicolumn{1}{c}{\textbf{\ldots}} \\  [-0.1em]
        \includegraphics[width=0.99\textwidth]{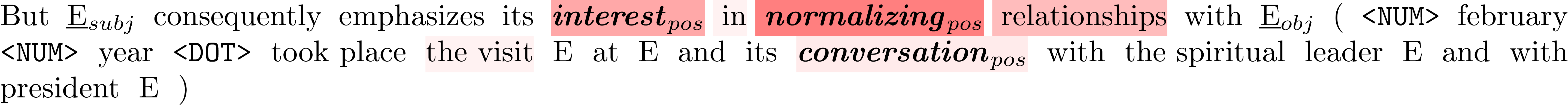} \\ \hline
    \end{tabularx}
    \captionof{figure}{
    Weight distribution visualization on sentiment contexts for model \modelCName{}, trained in different modes:
    supervised learning (SL), and with an application of distant supervision (DS);
    for visualization purposes,
    weight of each term is normalized by the maximum in context;
    frame entries (marked italic and bolded) appeared between masked attitude participants become greater weighted when
    training process employs \ruattitudes{} (DS mode).}
    \label{fig:heatmap}
\end{table*}

The terms quantification process remains a significant part of each attention-based encoder.
Being assigned and normalized, weights of every term in a context might be treated as
\textit{probability weight distribution} across all the terms appeared in a context.

The source of documents for contexts in this analysis is the \test{} part of the \rusentrel{} collection 
(Section~\ref{sec:evaluation}).
We analyse the weight distribution of the \framesGroup{} group,
declared in Section~\ref{sec:model}, across all input contexts.
We additionally introduce a list of extra groups utilized in the analysis by separating the subset of
\wordsGroup{} into
prepositions (\prepGroup{}),
terms appeared in \rusentilex{} lexicon (\sentGroup{}, Section~\ref{sec:data}),
nouns (\nounsGroup{}), and verbs (\verbsGroup{}).
The contents of \nounsGroup{} and \verbsGroup{} is considered only for those entries that are not present in the \rusentilex{} lexicon.

The \textit{context-level weight} of a particular term group
is a weighted sum of terms which both appear
in the context and belong to the corresponding term group.
For discrepancy analysis between sentiment and neutrally labeled contexts,
we utilize distributions of context-levels weights across:
\begin{enumerate}
\item \textbf{Sentiment contexts} (\textsc{S}) -- contexts, labeled with \textbf{positive or negative} labels;
\item \textbf{Neutral contexts} (\textsc{N}) -- contexts, labeled as \textbf{neutral}.
\end{enumerate}

Further, such weight distributions over sentiment and neutral contexts denoted as
\sD{\Any{}} and \nD{\Any{}} respectively, where asterisk corresponds to the certain term group.

To reveal the difference between distributions, the statistics
from Kolmogorov-Smirnov test was used~\cite{massey1951kolmogorov}.
In our analysis, the calculation of such statistics is considered to be performed between a pair of samples (tabulated distributions),
where each sample is a sequence of term group probabilities within each context.
It is worth to note that such tabulated distributions meet the criteria of the independence of values
(weights) related to \textit{continious} set.
Considering the latter, we are able to switch from tabulated to the
cumulative distributions as follows:

\newcommand{\cumD}[2]{\mathrm{F}^{#2}_{#1}}
\newcommand{\attSet}[1]{\mathrm{X}}

\begin{equation}
    \cumD{\attSet{}}{\Any{}}(x) = \mathrm{P}(\attSet{} < x) = \int_{-\infty}^x \rho^{\Any{}}_{\attSet{}}(t) dt
\end{equation}
where $\attSet{}$ is related to the contexts set of a certain polarity (sentiment or neutral), i.e.
$\attSet{} \in \{\textsc{S}, \textsc{N} \}, x \in [0, 1]$.
The Kolmogorov-Smirnov statistics (KS-statistics) represent the maximum
of the absolute deviation between cumulative distributions $\cumD{\textsc{S}}{\Any{}}$ and $\cumD{\textsc{N}}{\Any{}}$:
\begin{equation}
    D_{\Any{}} = \sup_{x \in [0, 1]}|\cumD{\textsc{S}}{\Any{}}(x) - \cumD{\textsc{N}}{\Any{}}(x)|
    \label{eq:D}
\end{equation}

Table~\ref{tab:ks_stat} provides the calculated KS-statistics (Formula~\ref{eq:D}) separately for each group of terms.
Larger values by $D_{\Any{}}$ address on a greater difference in weights distribution
between \sD{\Any{}} and \nD{\Any{}}.

Another statistics that we utilize in analysis is a difference in estimated values of \sD{\Any{}} and \nD{\Any{}}:
\begin{equation}
    \dd{\Any{}} = E(\sDm{\Any{}}) - E(\nDm{\Any{}})
    \label{eq:mean}
\end{equation}

In addition to KS-statistics, the calculation of $\dd{\Any{}}$ provides the sign of the difference.
Summarizing results of both statistics, we may conclude that
among all the models presented in our analysis,
only \bilstmPZhou{}
illustrates a significant difference
between \nD{} and \sD{}
across all the term groups.
The comparative kernel density estimations of context weight distributions for
\bilstmPZhou{} and \cnnEnds{}
is presented in Figure~\ref{fig:dist-analysis}.
In case of \bilstmPZhou{}, application of \ruattitudes{} in  training  (\textsc{DS} mode)
results in weights distribution biasing from
\nounsGroup{} and \prepGroup{}
onto terms of the \framesGroup{} and \sentGroup{} groups in sentiment contexts.
The similar case is observed for \cnnEnds{} trained in \textsc{DS} mode:
terms of \framesGroup{} and \sentGroup{} groups become more valuable equally in
sentiment and neutral context sets.
The assumption here is a structure of contexts in \ruattitudes{} (Section~\ref{sec:ruattitudes}):
all the contexts enriched with frames, appeared between attitude participants. 
Those cases where frames convey the presence of an attitude in context are presented in Figure~\ref{fig:heatmap}.
According to the provided examples for \bilstmPZhou{} model, it is possible to investigate greater weighted frame entries
when the training process of related model employs \ruattitudes{}.

Overall, the model \modelCName{} stands out baselines and models with feature-based attention encoders (\cnnEnds{}, \pcnnEnds{})
both due to results (Section~\ref{sec:experiments})
and the greatest discrepancy between \sD{} and \nD{}
across all the term groups presented in the analysis (Figure~\ref{fig:dist-analysis}).
We assume that the latter is achieved due to the following factors:
(1) application of bi-directional \lstm{} encoder;
(2)  utilization of a single trainable vector ($w$) in the quantification process
(Section~\ref{sec:self-based})
while the models of feature-based approach
(Section~\ref{sec:mlp-based}, Formula~\ref{eq:mlp_weight})
depend on fully-connected layers.

\section*{Conclusion}

In this paper, we study the attention-based models, aimed to extract sentiment attitudes from analytical articles.
We consider the problem of extraction as  two-class and three-class classification tasks for whole documents.
Depending on the task, the described models should classify a context with an attitude mentioned in it onto the
following classes: positive or negative (two-class); positive, negative, or neutral (three-class).

We investigated two types of attention embedding approaches: (1)~feature-based, (2)~self-based.
To fine-tune the attention mechanism, we utilized distant supervision technique by employing \ruattitudes{} collection in the training process.

We conducted experiments on Russian analytical texts of the \rusentrel{} corpus and provided  analysis of the results.
The affection of distant-supervision technique onto attention-based encoders
was shown by the variety in weight distribution of
certain term groups between sentiment and non-sentiment contexts.
Utilizing the distant-supervision approach in training three-class classification models results in
\dsBaseImproveRange{}\% improvement by \fmeasure{} for architectures that do not employ attention module in context encoder.
Replacing the latter with attention-based encoders provides the classification improvement by
\dsAttImproveRange{}\% \fmeasure{}.

In further work we plan to study application of language models for the presented tasks, as it continues the idea of attentive encoders application.

 \section*{Acknowledgments}
 The reported study was funded by RFBR according to the research project \textnumero~20-07-01059. 

\bibliographystyle{ACM-Reference-Format}
\bibliography{sample-authordraft}

\appendix

\end{document}